\pdfoutput=1

\documentclass[11pt]{article}

\usepackage{EACL2023}

\usepackage{times}
\usepackage{latexsym}
\usepackage{multirow}
\usepackage[T1]{fontenc}

\usepackage[utf8]{inputenc}

\usepackage{multirow}

\usepackage{microtype}
\usepackage[shortlabels]{enumitem}
\usepackage{caption}
\usepackage{subcaption}
\usepackage{graphicx}
\usepackage{booktabs}

\title{Cross-Lingual Transfer of Cognitive Processing Complexity}

\author{Charlotte Pouw \\ ILLC, University of Amsterdam\thanks{This research was developed when the first author was affiliated to Vrije Universiteit Amsterdam.} \\ \texttt{c.m.pouw@uva.nl}
        \And Nora Hollenstein \\ University of Copenhagen \\ \texttt{nora.hollenstein@hum.ku.dk}
        \AND Lisa Beinborn \\ Vrije Universiteit Amsterdam \\ \texttt{l.beinborn@vu.nl}}

\begin{document}\maketitle
\begin{abstract}
When humans read a text, their eye movements are influenced by the structural complexity of the input sentences. This cognitive phenomenon holds across languages and recent studies indicate that multilingual language models utilize structural similarities between languages to facilitate cross-lingual transfer. We use sentence-level eye-tracking patterns as a cognitive indicator for structural complexity and show that the multilingual model XLM-RoBERTa can successfully predict varied patterns for 13 typologically diverse languages, despite being fine-tuned only on English data. We quantify the sensitivity of the model to structural complexity and distinguish a range of complexity characteristics. Our results indicate that the model develops a meaningful bias towards sentence length but also integrates cross-lingual differences. We conduct a control experiment with randomized word order and find that the model seems to additionally capture more complex structural information.

\end{abstract}

\section{Introduction}
Approximately 7,000 languages are currently spoken in the world, exhibiting differences at almost every level of linguistic organization \cite{ethnologue}. Nonetheless, psycholinguistic theories are predominantly supported by evidence from a handful of Indo-European languages \cite{norcliffe-2015-crosslinguistic}. Only recently, researchers have started to explore cross-linguistic differences in the neural implementation of language, uncovering both striking similarities across languages and empirical differences that cannot be explained by a unitary account \cite{malik-moraleda-2022}.

In natural language processing, multilingual language models are optimized for tasks such as machine translation or cross-lingual information retrieval \cite{conneau-etal-2020-unsupervised} and follow a linguistically na\"{i}ve training regime. They are trained on dozens of languages simultaneously and do not account for typological differences between languages. Nevertheless, their cross-lingual transfer performance sets new records, even in zero-shot settings  \cite{pires-etal-2019-multilingual}. 
The ability to transfer knowledge across languages has been attributed to the shared vocabulary that is used for all languages \cite{wu-dredze-2019-beto} because it enables the reuse of common morphological roots for languages from the same family. 
However, recent studies indicate that vocabulary sharing is not a prerequisite for cross-lingual transfer \cite{artetxe-etal-2020-cross} and that structural commonalities between languages play a more prevalent role in models \cite{K2020Cross-Lingual}. 

Human sentence processing is sensitive to structural complexity. Eye movement data recorded during reading provide insights into cognitive processing patterns with a temporal accuracy of milliseconds \cite{winke-2013-eyetracking}. Structural processing difficulty materializes as regressions towards the complex region and an increase of fixations on that region \citep{clifton2011syntactic}. For example, sentences with an object-relative structure trigger more regressions than sentences with more common subject-relative clauses \cite{gordon-2006-similarity}. A classical example of structural complexity are garden-path sentences which initially trigger a simplified interpretation that must be revised when reading the rest of the sentence \citep{bever-1970-linguistic}. 

On the surface level, eye movement patterns are language-specific since they are influenced by visual factors such as orthography and word length \cite{kliegl-2004-length}. For example, the Chinese script is much more visually dense than the alphabetic script, resulting in longer fixations and saccades that move to positions relatively close to the current word \cite{LIVERSEDGE20161}. On a deeper processing level, reading patterns seem to converge across languages. Predictability effects have been demonstrated in multiple languages \cite{aljassmi2022effects,Laurinavichyute2019RussianSC} and sentences that are matched for content are read at a similar speed in Chinese, English, and Finnish \cite{LIVERSEDGE20161}.


\citet{sarti-etal-2021-looks} find that the representations of an English pre-trained transformer-based language model encode structural complexity more prominently when they are fine-tuned to predict English eye-tracking patterns. Interestingly, \citet{rama-etal-2020-probing} claim that structural similarity between languages is only weakly represented in multilingual models. Nevertheless, \citet{hollenstein-etal-2021-multilingual} show that multilingual models are able to predict eye movement patterns of reading even for languages that are not seen during fine-tuning, which indicates a general learnability of the relationship between structural complexity and eye movement patterns. Their results are restricted to four languages (three of them are from the Germanic family), and it remains unclear which structural cues are leveraged for the cross-lingual prediction because the test sentences are not aligned across languages. 

\paragraph{Contributions}
We examine whether the multilingual model XLM-RoBERTa (henceforth XLM-R) is sensitive to the structural complexity patterns that can be found in eye-tracking data. We use data from the newly released Multilingual Eye-tracking Corpus \citep{siegelman-2022-meco} to predict eye movement patterns for parallel texts in 13 typologically diverse languages. This allows us to specifically target the model's sensitivity towards structural information and rules out the possibility that the results are influenced by differences in semantics or dataset sizes.

We show that XLM-R can apply cross-lingual transfer to predict eye-tracking patterns for all 13 languages while being fine-tuned only on English eye-tracking data. Our results indicate that the model develops a meaningful bias towards sentence length, but also integrates cross-lingual differences. For a more detailed analysis of structural sensitivity, we probe the model's final layer for complexity features. Based on a control experiment with randomized word order, we conclude that the model seems to additionally capture more complex structural information.\footnote{\url{https://github.com/CharlottePouw/crosslingual-complexity-transfer}}

\section{Related Work} 
We introduce recent findings on the role of structural information for cross-lingual transfer in multilingual models and motivate the use of eye-tracking data as a proxy for cognitive processing complexity.

\subsection{Cross-lingual Transfer in Multilingual Models}
Massive multilingual language models such as mBERT \cite{devlin-etal-2019-bert} and XLM-R \cite{conneau-etal-2020-unsupervised} are trained on more than a hundred languages simultaneously. \newcite{wu-dredze-2019-beto} show that this approach leads to surprisingly strong performances in cross-lingual transfer settings and attribute the improvements to the shared subword vocabulary. \newcite{pires-etal-2019-multilingual} 
note that the model's ability to generalize "cannot be attributed solely to vocabulary memorization". Complementary, \newcite{artetxe-etal-2020-cross} and \newcite{liu2020study} find that a shared vocabulary is not necessary for cross-lingual transfer. Instead, the multilingual model seems to exploit structural similarity between the training and the target language to facilitate transfer \cite{K2020Cross-Lingual}.

Structural similarity is loosely defined as an overlap on a subset of typological characteristics which seem to be better reflected in multilingual language models explicitly optimizing for cross-lingual transfer  \cite{beinborn-choenni-2020-semantic,choenni2022investigating}. In language-agnostic models such as mBERT and XLM-R, the multilingual representations of the input can be separated into language-specific and language-neutral components \cite{tanti-etal-2021-language,libovicky-etal-2020-language,gonen-etal-2020-greek}. While \newcite{rama-etal-2020-probing} find that structural similarity between languages is only weakly represented in these models, \newcite{bjerva2019what} observe that structural similarity between languages correlates most with representational similarity. Experiments with artificial languages indicate that multilingual models are sensitive to hierarchical structure \cite{de-varda-zamparelli-2022-multilingualism} and to word order \cite{chai-etal-2022-cross,deshpande-etal-2022-bert}. 
\newcite{ahmad-etal-2021-syntax} show that cross-lingual transfer can be improved by explicitly encoding structural information via an auxiliary syntactic objective and \newcite{guarasci-2022-bert} find that structural complexity knowledge can even be transferred across languages without explicit training. 



\subsection{Predicting Processing Complexity}
Recent studies indicate that transformer-based language models are sensitive to structural characteristics of the input sentence when predicting eye-tracking patterns. \citet{hollenstein-etal-2021-multilingual} find a correlation between the Flesch reading ease score and eye-tracking prediction accuracy of pre-trained multilingual transformer models which disappears after fine-tuning. \citet{wiechmann-kerz-2022-measuring} detect similar correlations between the prediction accuracy of English transformer models and a wider range of readability features. Finally, \citet{hollenstein-etal-2022-patterns} find that eye-tracking metrics predicted by multilingual transformer models correlate in a similar way with readability features as eye-tracking metrics recorded from human readers.

Sensitivity to structural complexity also seems to increase when incorporating eye-tracking data in NLP models. Learning eye movement behavior as an auxiliary task has been shown to facilitate the prediction of text complexity in English and Portuguese \cite{gonzalez-garduno-sogaard-2017-using,evaldo-leal-etal-2020-using}. \citet{barrett-etal-2016-cross} show that English eye-tracking features improve the performance a French part-of-speech tagger, suggesting that information learned from monolingual eye-tracking data is transferable across languages.

In this work, we explicitly test for sensitivity to a range of structural characteristics in multilingual models and analyze if structural sensitivity increases by learning to predict eye-tracking patterns. We extend previous analyses to a much wider range of languages from five different families (Indo-European, Koreanic, Semitic, Turkic, and Uralic).

\section{Methodology}
We fine-tune a pre-trained multilingual transformer model to predict eye-tracking metrics in a setting of zero-shot cross-lingual transfer. 

\subsection{Data}
We use the aligned multilingual eye-tracking corpus MECO for testing. As the multilingual data consists of only few samples, we use the larger monolingual English eye-tracking dataset GECO for training. Size statistics of both corpora can be found in the appendix in Table \ref{tab:statistics}.

\paragraph{Multilingual Eye-tracking Corpus (MECO)}
The Multilingual Eye-tracking Corpus contains parallel eye-tracking data of reading in 13 different languages \cite{siegelman-2022-meco}.\footnote{Dutch, English, Estonian, Finnish, German, Greek, Hebrew, Italian, Korean, Norwegian, Russian, Spanish, Turkish.} The reading material consists of 12 short Wikipedia-style texts about various topics, which participants read in their native language. The texts were either directly translated or carefully matched for topic, genre, and readability. Each of the 12 texts was presented on a single screen and in the same fixed order in all languages. The number of participants ranged from 29 to 54 per language (45 on average).

\paragraph{Ghent Eye-tracking Corpus (GECO)}
The Ghent Eye-tracking Corpus contains eye-tracking data from 14 monolingual English readers \cite{cop-2016-geco}. They were reading the entire novel \textit{The Mysterious Affair at Styles} by Agatha Christie which was presented on the screen one paragraph at a time. 

\subsection{Experimental Setup}
We use multi-task learning for predicting four sentence-level eye-tracking metrics. 

\paragraph{Sentence-Level Eye-Tracking Metrics}
\citet{LIVERSEDGE20161} find that eye movement patterns are more comparable across languages at the sentence level than at the word level.
We select four sentence-level eye-tracking metrics that cover both early and late language processing in line with \citet{sarti-etal-2021-looks}. For each sentence \textit{s}, we consider:

\begin{enumerate}[noitemsep]
    \item \textit{Fixation count}: number of fixations on \textit{s}
    \item \textit{Total fixation duration}: total duration of all fixations on \textit{s}
        \item \textit{First-pass duration}: duration of the first reading pass over \textit{s}
    \item \textit{Regression duration}: total duration of all regressions within \textit{s}.
\end{enumerate}
Duration values are measured in milliseconds. To obtain generalized eye movement patterns, we average all eye-tracking metrics over participants and scale each eye-tracking feature to fall in the range 0--100, so that the loss can be calculated uniformly for durations and counts \citep{hollenstein-etal-2021-multilingual}. The distribution of the four metrics is shown in the appendix in Figure \ref{fig:distribution-geco-meco}.


\begin{figure*}
    \centering
    \includegraphics[scale=0.47]{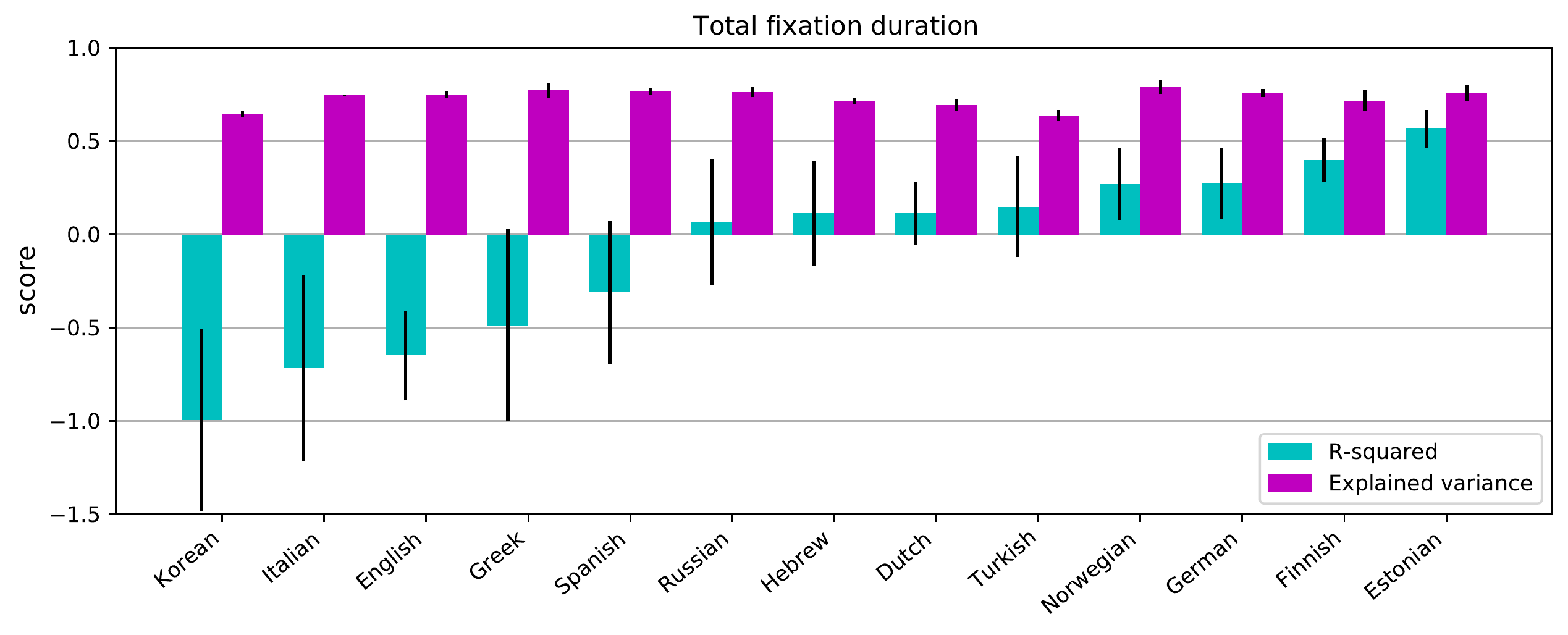}
    \caption{Cross-lingual transfer results for predicting cognitive processing complexity (i.e. sentence-level fixation duration). Prediction performance is evaluated with explained variance and $R^{2}$ for each language in MECO. The results are averaged over 5 folds; error bars denote the standard deviation over folds.}
    \label{crosslingual-accuracy-tfd}
\end{figure*}
\paragraph{Model}\label{model}
We use XLM-R \citep{conneau-etal-2020-unsupervised} as our multilingual transformer model since it achieved the best zero-shot results in the CMCL 2022 Shared Task on Multilingual and Crosslingual Prediction of Human Reading Behaviour \cite{srivastava-2022-poirot,hollenstein-etal-2022-cmcl}. The model was pre-trained on 2.5TB CommonCrawl data containing 100 languages using the Masked Language Modelling objective and uses SentencePiece subword tokenization \cite{kudo-richardson-2018-sentencepiece}. We select the Huggingface checkpoint \textit{xlm-roberta-base} and add a linear dense layer to predict four sentence-level eye-tracking metrics. 

\paragraph{Multi-Task Learning}
We employ multi-task learning with hard parameter sharing to fine-tune the model on all eye-tracking metrics simultaneously in line with \newcite{sarti-etal-2021-looks}. This means that all model parameters are shared except for the task-specific regression heads in the final prediction layer. More specifically, the same sentence representation is fed into each of the four regression heads which predict their respective eye-tracking metric. The model parameters are optimized jointly for all regression tasks by summing the individual MSE losses in line with previous work \cite{hollenstein-etal-2021-multilingual,hollenstein-etal-2022-cmcl,wiechmann-kerz-2022-measuring}.

\paragraph{Training Parameters} We fine-tune XLM-R for 15 epochs with early stopping after 5 epochs without an improvement in the validation accuracy. We use 10\% of the training data as validation data and evaluate every 40 steps. We employ a batch size of 32 and a learning rate of 1e-5. The sentence representation is obtained by mean pooling over token representations. We train the model on the GECO data using 5-fold cross-validation and report the average over the folds for each language in MECO.


\paragraph{Evaluation}\label{evaluation}
We report explained variance and R-Squared ($R^{2}$) to capture the proportion of variance in the dependent variable that can be explained by our model in line with \citet{sarti-etal-2021-looks}. Explained variance uses the biased variance to determine what fraction of the variance is explained. $R^{2}$ uses the raw sums of squares instead and provides complementary information about systematic offsets in the predictions. We report both metrics and evaluate the performance of the fine-tuned model individually for each of the four eye-tracking metrics.\footnote{In previous work on token-level eye-tracking prediction, the mean absolute error was reported instead but it is less informative for sentence-level predictions because sentence-level eye-tracking metrics are generally more centered around the mean.}



\section{Cross-Lingual Transfer Results}\label{xling}

Figure \ref{crosslingual-accuracy-tfd} shows the explained variance and $R^{2}$ scores of the fine-tuned model for total fixation duration across languages. In terms of explained variance, we see that the model achieves a similar performance across languages, i.e. it captures 60 to 80 percent of the variance in the original eye-tracking signal for all languages. The $R^{2}$ scores, on the other hand, vary much more depending on the language. Similar results were observed for two of the other eye-tracking metrics, i.e. fixation count and first-pass duration, but the model is worse at predicting regression duration (see Figure \ref{crosslingual-accuracy-appendix} in the appendix). To better control for spurious correlations, we ran the experiment on permuted input-output pairs, i.e., we paired input sentences with eye-tracking values corresponding to another random sentence and averaged the results over 5 folds. For this random baseline setup, both explained variance and $R^{2}$ are always strictly negative for all languages. 

\begin{figure*}[h!]
\begin{subfigure}{.5\textwidth}
  \centering
  \includegraphics[width=.9\linewidth]{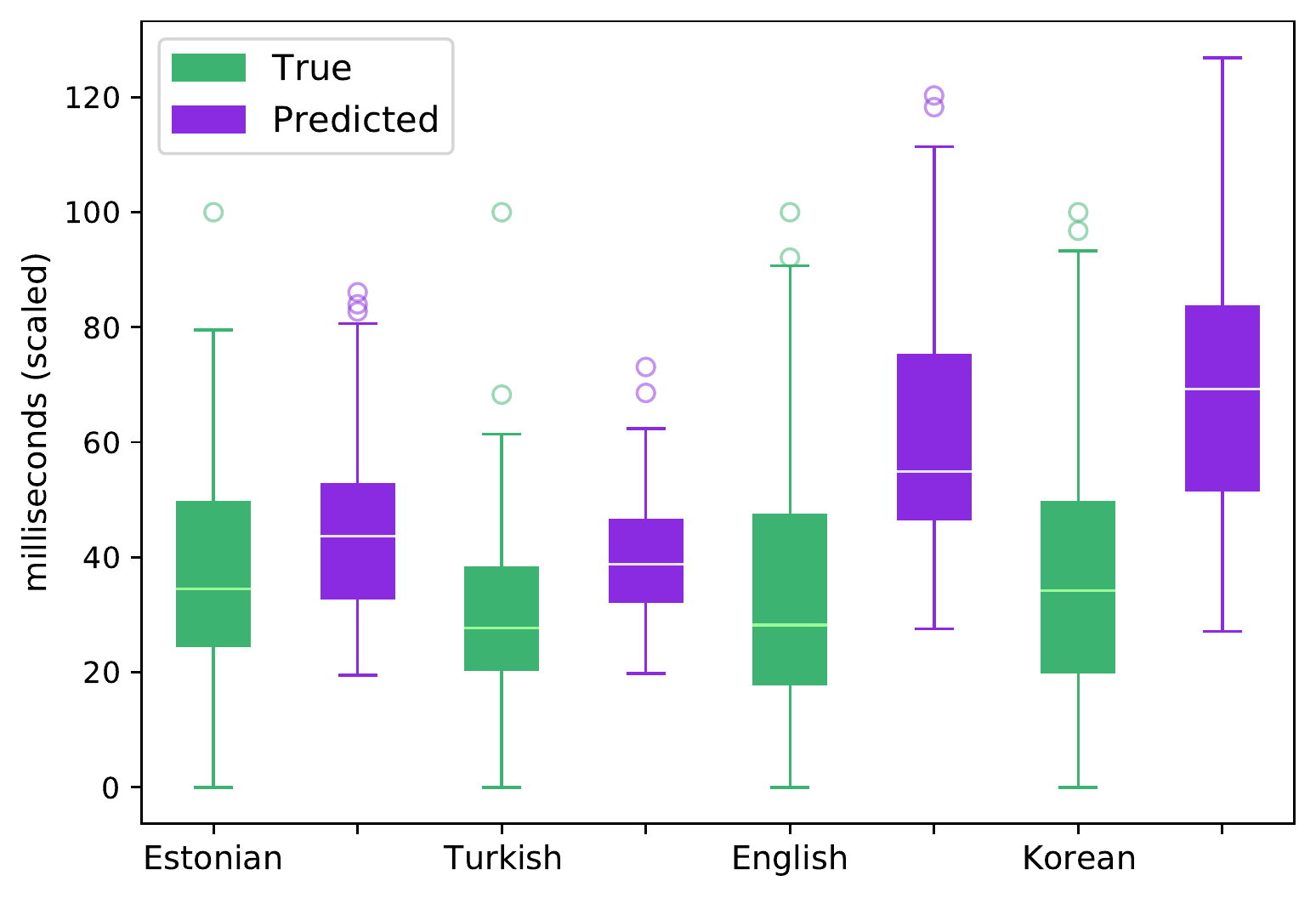}
  \label{fig:sub1}
\end{subfigure}%
\begin{subfigure}{.5\textwidth}
  \centering
  \includegraphics[width=.9\linewidth]{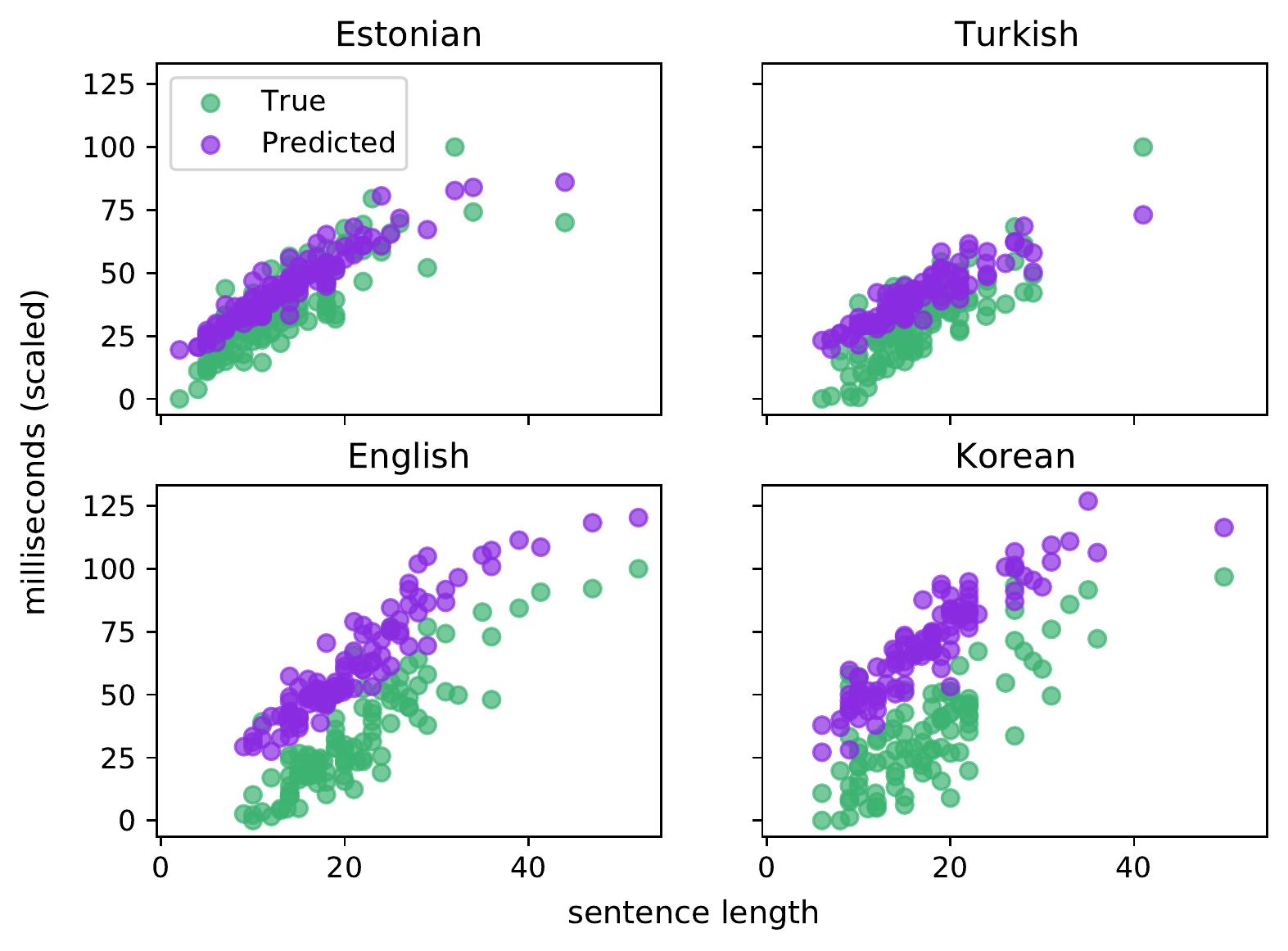}
  \label{fig:sub2}
\end{subfigure}
\caption{The left plot shows the distribution of true and predicted values for total fixation duration for Estonian, Turkish, English and Korean sentences in MECO. The right figure shows the distribution of values with respect to sentence length.}
\label{fig:true-vs-pred}
\end{figure*}

To better understand the varied $R^{2}$ scores for different languages, we show the distribution of the true and predicted values for total fixation duration for two languages with high $R^{2}$ (Estonian, Turkish) and two languages with low $R^{2}$ (English, Korean) in Figure \ref{fig:true-vs-pred}. We see that the low $R^{2}$ for English and Korean is caused by predictions that are consistently too high. For Estonian and Turkish, the difference between true and predicted values is clearly smaller, resulting in a higher $R^{2}$. Nevertheless, the model is able to predict a significant amount of the variance in the eye-tracking signal of all languages, as expressed by the stable explained variance scores across languages.

Interestingly, the model performs slightly better for most zero-shot languages than for the fine-tuning language English. Recall that this performance difference cannot be attributed to cross-lingual differences in semantics, since all sentences are parallel with respect to content. On the right side of Figure \ref{fig:true-vs-pred}, we analyze the predictions with respect to sentence length and find that both the model predictions and the true values for fixation duration correlate with sentence length in all languages. As sentence length is an indicator of structural complexity, we further dissect this phenomenon and conduct an analysis of a range of structural characteristics in the following section.



\section{Sensitivity to Structural Complexity}
We explore four categories of sentence-level complexity features: length, frequency, morpho-syntactic, and syntactic. Word frequencies are obtained as standardized Zipf frequencies using the Python package wordfreq \citep{robyn_speer_2018_1443582}. The package combines several frequency resources, including SUBTLEX lists (e.g. \citet{brysbaert-2009-english}) and OpenSubtitles \citep{lison-tiedemann-2016-opensubtitles2016}. The morpho-syntactic and syntactic features are computed using the Profiling-UD tool \citep{brunato-etal-2020-profiling}.  

\begin{table*}[h!]
\centering
\small

\begin{tabular}{llcccc}
\toprule
\textbf{Example}&&&&&Prediction\\
\midrule
English & \multicolumn{4}{l}{\textit{In ancient Roman religion and myth, Janus is the god of beginnings and gates.}}& \textbf{42.96}\\
Finnish & \multicolumn{4}{l}{\textit{Muinaisen roomalaisen mytologian mukaan Janus oli alkujen ja porttien jumala.}}& 38.91 \\
Turkish & \multicolumn{4}{l}{\textit{Antik Roma inan{\i}{\c{s}}lar{\i}nda ve mitlerinde, Janus ba{\c{s}}lang{\i}{\c{c}}lar{\i}n ve kap{\i}lar{\i}n tanr{\i}s{\i}d{\i}r.}} & 32.28 \\
\midrule
\vspace{-1em}
\end{tabular}
\begin{tabular}{llllll}
\midrule
\textbf{Structural Complexity}&             & English        & Finnish       & Turkish &  \\ \midrule
Length&Sentence length (tokens)             & \textbf{14}    & 10            & 10    &    \\
&Avg. word length (characters)            & 4.57           & 6.80          & \textbf{7.60}&\\

Frequency&Avg. word frequency (Zipf)         & 5.63           & 4.36          & \textbf{3.46}& \\
&\# low frequency words      & 2              & \textbf{6}    & \textbf{6} &  \\

Morpho-Syntactic&Lexical density             & 0.57           & 0.70          & \textbf{0.73}   &       \\

Syntactic&Parse tree depth            & 3              & 3             & 3    &        \\
&Avg. dependency link length       & 2.15           & \textbf{2.78} & 1.90    &      \\
&Max. dependency link length       & \textbf{7}     & \textbf{7}    & 4      &       \\
&\# verbal heads             & 1              & 1             & 1      &       \\ \bottomrule
\end{tabular}
\caption{Predicted values for total fixation duration for the same example sentence in English, Finnish, and Turkish (top), and the respective values for the nine structural complexity features (bottom).}
\label{tab:complexity-features}
\end{table*}


\paragraph{Cross-Lingual Differences}\label{cross-lingual-differences}
We showcase an individual example sentence in Table \ref{tab:complexity-features} to compare the predicted fixation duration for English, Finnish and Turkish. We observe that the highest value is predicted for the English version. This is most likely caused by its length, as the sentence is less complex than the Finnish and Turkish versions in terms of all other linguistic features. 

Interestingly, the model predicts that Finnish readers will fixate on the sentence longer than Turkish readers, even though both sentences have the same length. The Turkish sentence contains longer, less frequent words, and is lexically more dense, but the Finnish sentence contains longer dependency links. This indicates that the model is more sensitive to dependency structure than to low-level complexity (i.e. word length and frequency) when predicting eye-tracking values for sentences of the same length. 


\subsection{Sensitivity to Fine-Tuning Input}\label{compAnalysis}

To analyze the model's sensitivity to the structural complexity of the fine-tuning data, we compare the performance of the fine-tuned model for in-domain data (English GECO) and cross-domain data (English MECO). Table \ref{cross-domain} shows the explained variance and $R^{2}$ scores of the fine-tuned model predictions for each eye-tracking metric for both domains. We see that the model consistently yields more accurate predictions for the in-domain data than for the cross-domain data. 

\begin{table}[h!]
\centering
\small
\begin{tabular}{@{}lrrrr@{}}
\toprule                    & \textbf{MECO} &       & \textbf{GECO} &  \\ \midrule
                        & EV        & $R^{2}$         & EV       & $R^{2}$     \\
                        \midrule
\textbf{FC}          &  .78 (.02)     &  -.63 (.35)     &   .93 (.00)   & .93 (.01)      \\
\textbf{TFD} &  .75 (.02)     &  -.65 (.24)    &    .92 (.00)  &  .92 (.01) \\
\textbf{FPD}    &  .50 (.03)   &  -.87 (.27)    &    .95 (.00)   & .95 (.01)       \\
\textbf{RD}     &  -.28 (.14)   &  -.96 (.45)    &    .44 (.04)   &  .45 (.05)     \\ \bottomrule
\end{tabular}
\caption{Explained variance (EV) and $R^{2}$-scores of the fine-tuned model predictions for four eye-tracking metrics from the English parts of MECO and GECO: fixation count (FC), total fixation duration (TFD), first-pass duration (FPD), and regression duration (RD). The results are averaged over 5 folds; standard deviations are indicated in parentheses.}
\label{cross-domain}
\end{table}



\begin{figure*}
\centerline{\includegraphics[scale=0.47]{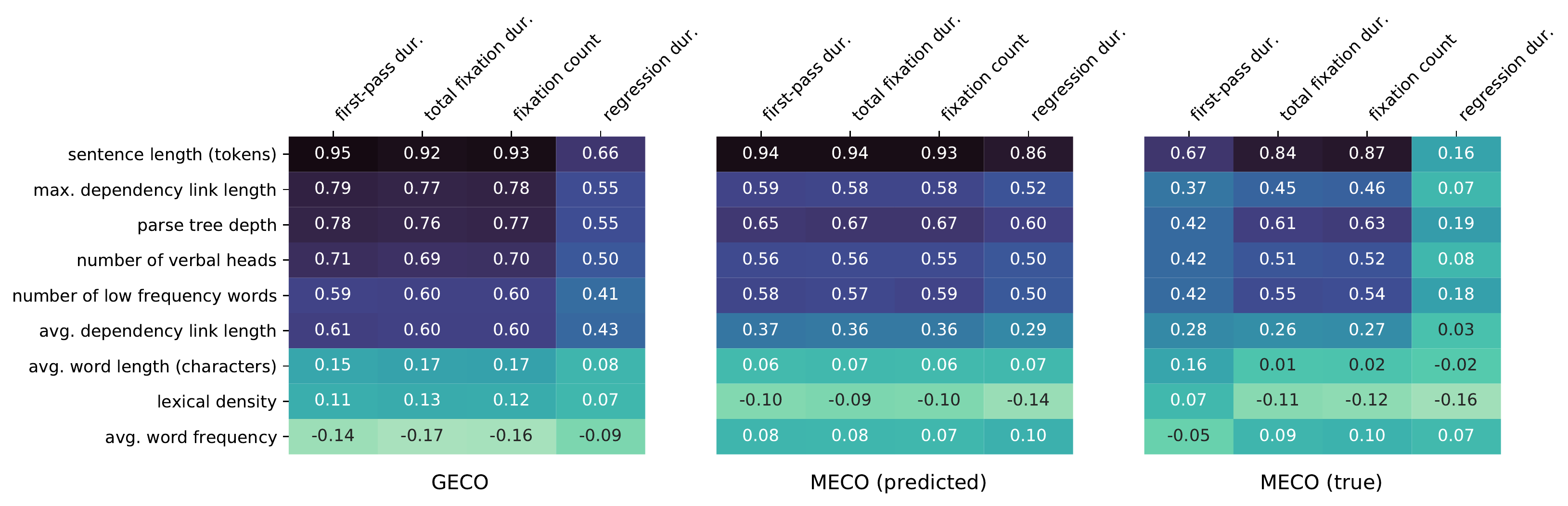}}
\caption{Spearman correlations between complexity features and eye-tracking metrics of GECO and the English part of MECO (predicted versus true). A darker color represents a stronger correlation. All GECO correlations are significant (\textit{p} $<$ 0.001); MECO correlations above 0.2 are significant (\textit{p} $<$ 0.01).}
\label{fig:correlations}
\end{figure*}

To better understand why the model does not generalize well across domains for English, we visualize the Spearman correlation between complexity features and eye-tracking metrics for English GECO and MECO sentences in Figure \ref{fig:correlations}. We see that the predicted values for the MECO sentences exhibit a similar correlation pattern with the complexity features as the GECO sentences. The true values of MECO are less consistent with this pattern. Literary texts contain very different words than encyclopedic texts, which might influence fixation durations and trigger regressions that cannot solely be explained by structural complexity. In addition, MECO is significantly smaller than GECO (99 vs 4,041 English sentences) and contains data from a higher number of participants (46 vs 14). The smaller amount of sentences and the larger amount of readers increase the effect of individual differences\footnote{A higher number of participants leads to more diversity across readers with respect to individual factors that could influence reading strategies (e.g. age, education level). The GECO data came from 14 English readers who were all undergraduate students with an age range of 18-26. The MECO data came from 29 to 54 readers per language (45 on average), who had more diverse educational backgrounds and a wider age range (18-45). Based on these statistics, we assume that the increased heterogeneity of the MECO participants influences the correlations observed in Figure 3.} which might obscure correlations between structural complexity and eye movement patterns. Directly applying the learned correlations from GECO to MECO might explain why the fine-tuned model fails to generalize across domains.

The average sentence length is considerably higher in GECO than in MECO (21 vs 13 words, see Table \ref{tab:statistics}). As the model predictions strongly correlate with sentence length, we speculate that the model overestimates eye-tracking values for sentences that are longer than the majority of fine-tuning sentences which would explain the higher mean of the predictions in Figure \ref{fig:true-vs-pred}.

\paragraph{Multi-Task Learning Effect} Figure \ref{fig:correlations} further shows that regression duration is only weakly correlated with the complexity metrics in contrast to the other eye-tracking metrics. Nevertheless, the correlations between the model predictions and the complexity features are similar for all four metrics. This indicates a drawback of multi-task learning: since the loss is computed jointly over all tasks, accurate predictions for three out of four tasks already yield a small loss. The model seems to overfit to first-pass duration, total fixation duration and fixation count, which can all be predicted from similar complexity features, and does not learn the deviating patterns to predict regression duration. Further research is needed to better understand the linguistic features underlying regression duration.


\subsection{Feature-Based Prediction}

To further establish which complexity features are good predictors for each individual eye-tracking metric, we examine the extent to which the four eye-tracking metrics can be predicted from explicit features. Since multi-task learning seems to have a negative impact on learning the structural features underlying each individual eye-tracking metric, we train a separate feature-based model for each eye-tracking metric individually. We use support vector machines (SVM) with a linear kernel as our feature-based regression models. We employ the SVR implementation from scikit-learn \citep{scikit-learn} with all default parameters and use different subsets of features from Table \ref{tab:complexity-features}: 1) only the two length features, 2) only the two frequency features, 3) only the five structural (i.e., morpho-syntactic and syntactic) features, and 4) all nine features.

\begin{figure}[h!]
\centerline{\includegraphics[scale=0.35]{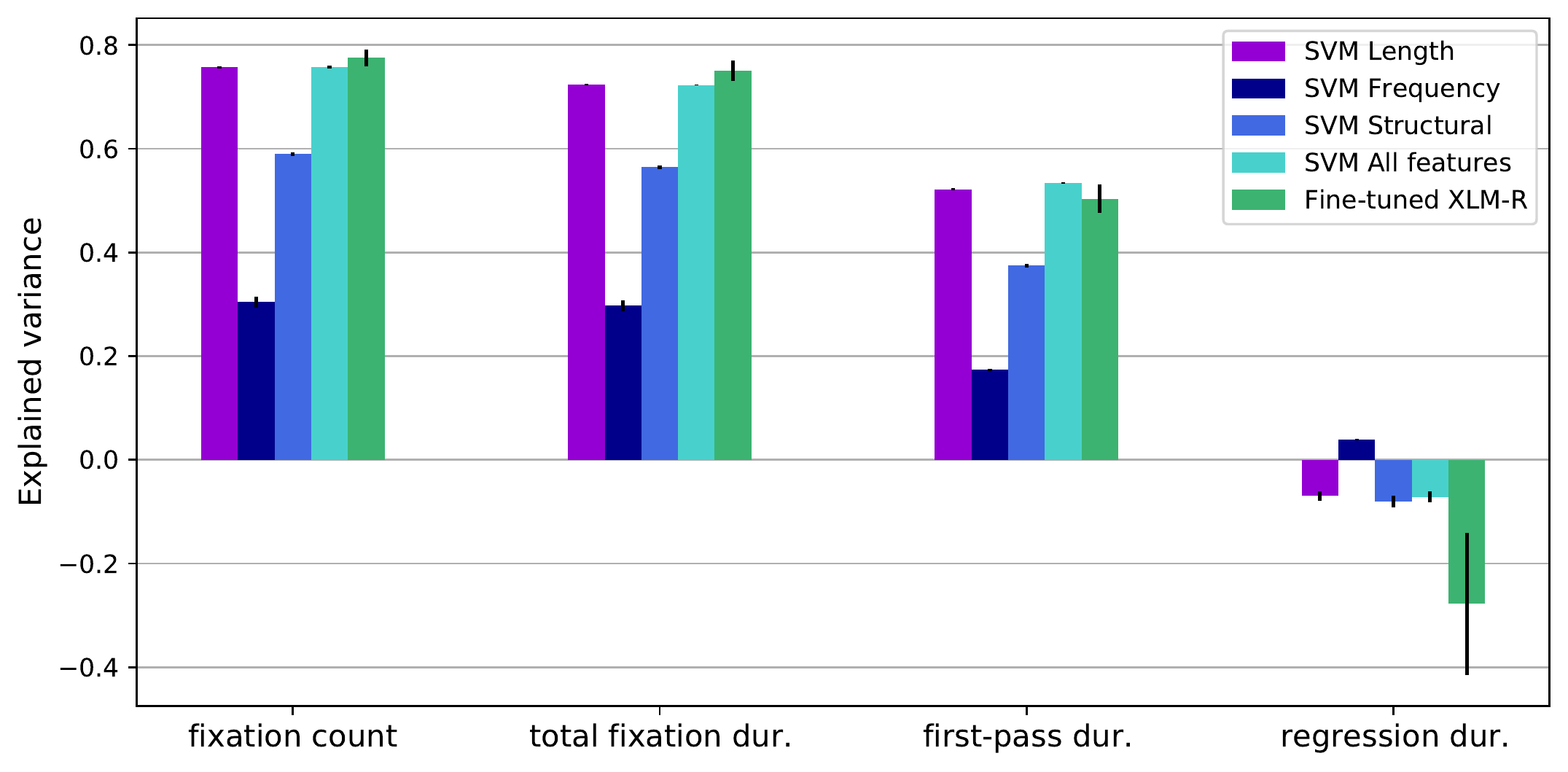}}
\caption{Explained variance of the four feature-based SVM models and the fine-tuned XLM-R model. The models are trained on GECO using 5-fold cross-validation and evaluated on the English part of MECO; error bars denote the standard deviation over folds.
}
\label{fig:feature-based-prediction-ev}
\end{figure}

As the SVM models predict a simpler problem (a single eye-tracking metric), it is not surprising that they outperform the fine-tuned multi-task model with respect to the absolute predictions (as measured by $R^2$, see appendix Figure \ref{fig:feature-based-prediction-r2}). More interestingly, Figure \ref{fig:feature-based-prediction-ev} shows that the multi-task model is able to capture a similar amount of variance as the length-based SVM. Furthermore, we see that the length-based SVM performs almost identically to the SVM trained on \textit{all} complexity features, outperforming the SVMs trained on frequency features and structural features. This shows that length is a strong predictor for sentence-level eye-tracking metrics, and suggests that structural and frequency features do not provide much additional information. We further investigate if length is the main factor affecting the predictions of the fine-tuned model in the following section.

\section{The Role of Sentence Length}
To test whether the fine-tuned XLM-R model captures more sophisticated structural information than sentence length, we conduct two additional experiments. First, we probe the final-layer representations of the model for the complexity features from Table \ref{tab:complexity-features}, both before and after fine-tuning on eye-tracking data. Second, we compare the performance of the fine-tuned model to a control condition: we randomize the word order within each MECO sentence to analyze the prediction performance on scrambled input.

\subsection{Probing Set-up}
We train regressors $g_i$ to predict a value for each of the nine latent factors of structural complexity $Z = z_1$,...,$z_9$ using XLM-R's final-layer representation $\theta(x)$ of our input sentence $x$. The prediction accuracy of $g_i$ is an indication of how prominently the linguistic property $z_i$ is encoded in $\theta$. We analyze this both for the pre-trained and fine-tuned representations of XLM-R to quantify the relative increase of sensitivity to $z_i$ after fine-tuning on eye-tracking metrics.

We conduct the probing experiments for three typologically different languages to analyze if the structural sensitivity that was acquired from English eye-tracking data transfers to other languages. As input, we use 1,000 parallel sentences from the English, Korean and Turkish parts of the Parallel Universal Dependencies (PUD) treebanks which were randomly selected from Wikipedia and news articles \cite{zeman-etal-2017-conll}. We apply a 5-fold cross-validation setting with 800 sentences for training the probing regressors for each language and the remaining 200 for testing. 
We use the same architecture as described in Section~\ref{model}, but freeze the encoder model and only update the final regression layer during training. The regression layer contains nine probing heads (one for each linguistic feature) and is trained for 5 epochs.\footnote{We report results for a multi-task set-up for probing in line with \citet{sarti-etal-2021-looks} and use the same hyperparameters as for the fine-tuning experiments but without intermediate evaluation on a development set. We also ran single-task probing as a sanity check and obtained similar results.}

\subsection{Results}

We report the results of the probing experiments and the model performance on scrambled inputs.

\paragraph{Probing}

\begin{figure}[h!]
    \centering
    \includegraphics[scale=0.35]{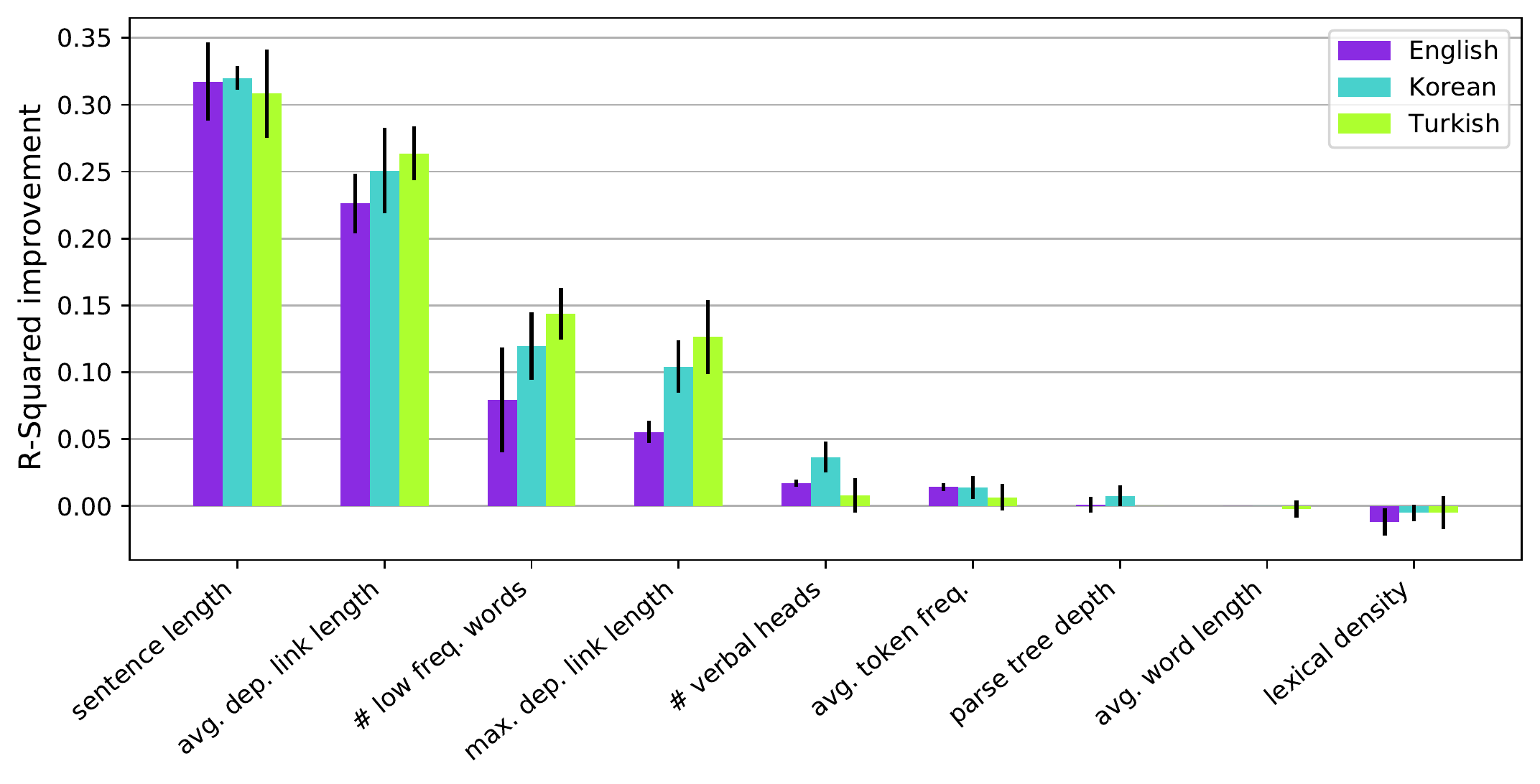}
    \caption{Relative improvement in $R^{2}$ for complexity features of English, Korean and Turkish sentences in fine-tuned XLM-R sentence representations over pre-trained representations. The results are calculated using probing regressors and averaged over 5 folds.} 
    \label{probing-results}
\end{figure}

Figure \ref{probing-results} shows the relative probing performance for each complexity feature. We see that fine-tuning yields the largest improvements for probing sentence length and average dependency link length. For the other complexity features, we see that the fine-tuned representations yield little to no improvement in probing accuracy compared to the pre-trained representations. This mostly concerns the features for which sentence length is factored out, i.e., average word frequency, average word length and lexical density. \citet{sarti-etal-2021-looks} report similar results and show that increased probing performance for dependency features persists for sentences of the same length. This provides additional evidence that structural information is learned in addition to low-level length information.

We observe only minor differences in probing accuracy for individual complexity features of English, Korean and Turkish sentences. The general pattern is consistent for all languages: features related to the structural complexity of sentences are more easily predicted after fine-tuning on eye-tracking metrics. This indicates that the fine-tuned model is able to transfer structural complexity knowledge acquired from English eye-tracking data to other languages.

\paragraph{Influence of Word Order} 

We compare the performance of the fine-tuned model on sentences with normal versus scrambled word order, both in terms of explained variance and $R^{2}$. We measure similar explained variance scores for both input types. This indicates that the model is able to account for a large portion of the variance in our eye-tracking data by merely considering sentence length. The $R^{2}$ scores, on the other hand, are consistently lower for scrambled inputs, as shown for total fixation duration in Figure \ref{fig:scrambled_results} (see appendix Figure \ref{scrambled-results-appendix} for the other eye-tracking metrics). We conclude that the model is sensitive to word order and bases its eye-tracking predictions not only on sentence length but also on more complex structural characteristics.

\begin{figure}[h!]
    \centering
    \includegraphics[scale=0.35]{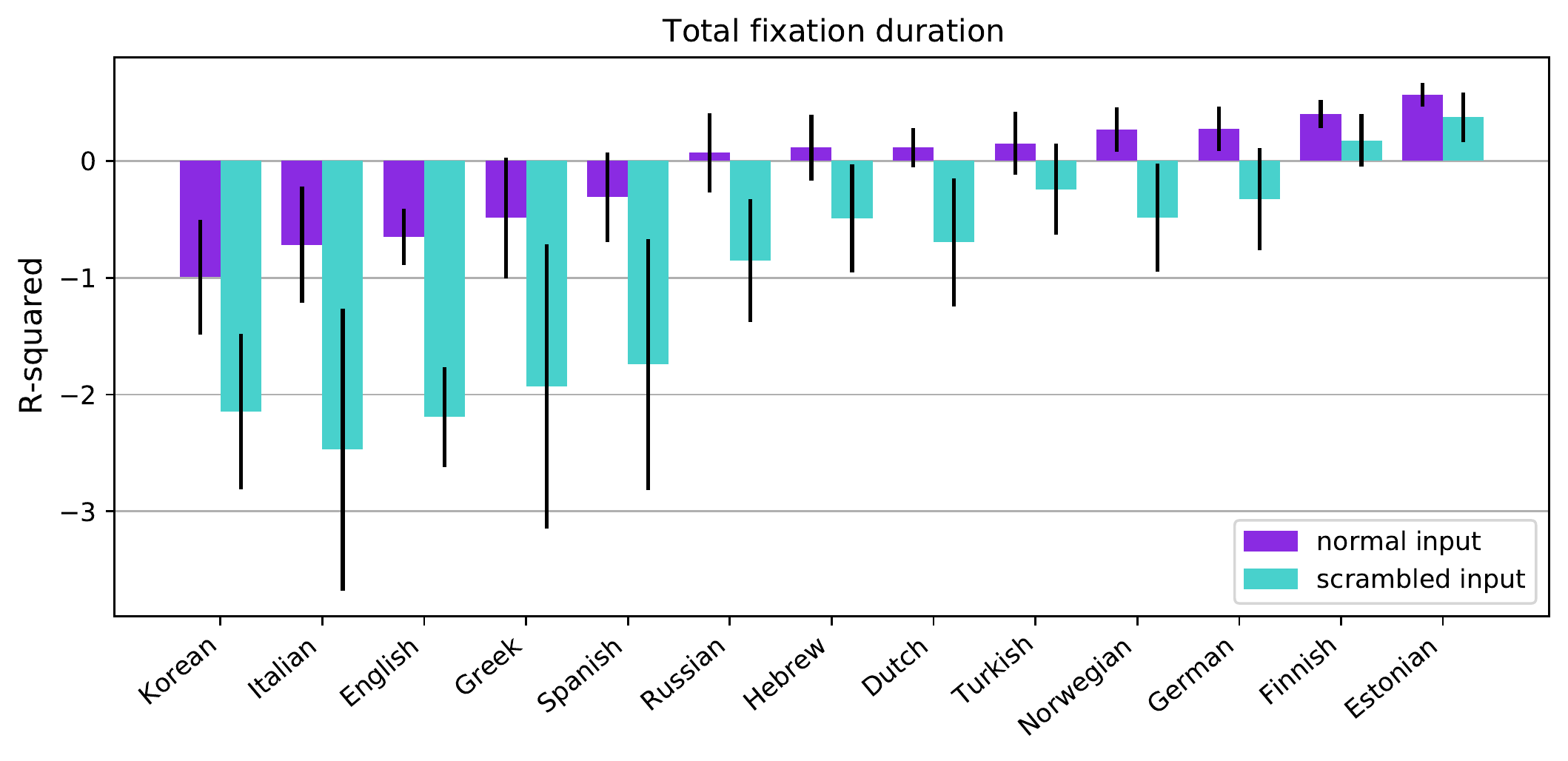}
    \caption{$R^{2}$ scores for total fixation duration for each language in MECO, both for sentences with normal and scrambled word order. The results are averaged over 5 folds; error bars denote the standard deviation.} 
    \label{fig:scrambled_results}
\end{figure}

\section{Conclusion}

We find that XLM-R can apply cross-lingual transfer to predict cognitive processing difficulty with similar performance across 13 typologically diverse languages, despite being fine-tuned only on English data. We conducted a range of experiments to quantify the model's sensitivity to structural complexity and find that the fine-tuned model prominently encodes sentence length, but also considers more complex structural information such as dependency structure and word order for the prediction of eye-tracking metrics.

Our analyses suggest that domain differences in training and testing data have a greater impact on model performance than language differences within the same domain. More specifically, XLM-R performs better on in-domain GECO data than cross-domain MECO data, but within MECO, XLM-R shows similar performance across languages. This aligns with the findings of \citet{morger-etal-2022-cross}, who show that the correlation between relative importance metrics and total fixation duration is influenced by text domain. Our study highlights the significance of controlling for text domain and size, as it allows to evaluate cross-lingual generalization that is independent of dataset characteristics.

In future work, we plan to better account for individual differences between readers \cite{brandl-hollenstein-2022-every} and spill-over effects across sentence boundaries \cite{wiechmann-kerz-2022-measuring}. The modeling approach for learning eye-tracking patterns also needs further exploration. 
We find that sentence-level prediction of eye-tracking patterns works well for learning about structural complexity, but that it is not optimal for capturing lexical complexity. Token-level measures, as predicted in \newcite{hollenstein-etal-2021-multilingual}, are more likely to be informative about lexical phenomena. A joint loss for sentence and token-level eye-tracking metrics might lead to sensitivity to a wider range of linguistic complexity features.


\section{Limitations}

The main limitation of our work is the use of relatively small datasets for testing our models due to limited availability of eye-tracking data in multiple languages. The dataset used for testing cross-lingual transfer (MECO) contains approximately 100 sentences per language. For probing structural complexity, we used a sample of 1,000 sentences per language.  

As in related work, we averaged the eye-tracking metrics over readers to obtain a more robust indication of human reading behavior. This approach disregards the fact that reading is a highly individual process that is dependent on cognitive factors and experience. A computational model might develop a better sense of linguistic complexity when it learns about the linguistic properties that lead to variation across readers and we are working towards methods for integrating this information.

\section*{Acknowledgements}
We thank the anonymous reviewers for their insightful feedback.
L. Beinborn's research was supported by the Dutch National Science Organisation (NWO) through the projects  CLARIAHPLUS (CP-W6-19-005) and VENI (Vl.Veni.211C.039).

\bibliography{anthology,custom}
\bibliographystyle{acl_natbib}

\clearpage
\appendix
\onecolumn
\section{Additional Tables and Figures}
\begin{table}[h!]
\centering
\small
\begin{tabular}{@{}ll|rrrr@{}}
\toprule
\textbf{Dataset} & \textbf{Language}  & \#\textbf{Words} & \#\textbf{Sentences} & \textbf{Avg. sent. length} & \textbf{Avg. word length} \\ \midrule
GECO    & English   & 52131   & 4041        & 12.90                & 4.60             \\ \midrule
MECO    & English   & 2092    & 99          & 21.13                & 5.32             \\
        & Dutch     & 2226    & 112         & 19.88                & 5.54             \\
        & German    & 2019    & 115         & 17.56                & 6.38             \\
        & Finnish   & 1462    & 110         & 13.29                & 8.19             \\
        & Estonian  & 1542    & 112         & 13.77                & 7.35             \\
        & Norwegian & 2106    & 116         & 18.16                & 5.62             \\
        & Italian   & 2111    & 90          & 23.46                & 5.70             \\
        & Spanish   & 2412    & 98          & 24.61                & 5.01             \\
        & Greek     & 2082    & 99          & 21.03                & 5.67             \\
        & Turkish   & 1696    & 104         & 16.31                & 6.92             \\
        & Russian   & 1827    & 101         & 18.09                & 6.53             \\
        & Hebrew    & 1943    & 121         & 16.06                & 4.89             \\
        & Korean    & 1699    & 101         & 16.82                & 3.21             \\
                                                \bottomrule
\end{tabular}
\caption{Size characteristics for the reading materials of GECO and MECO. GECO sentences which are shorter than five words are removed to ensure that the model sees an adequate amount of complex structures during training.}
\label{tab:statistics}
\end{table}

\begin{figure}[h!]
\centerline{\includegraphics[scale=0.5]{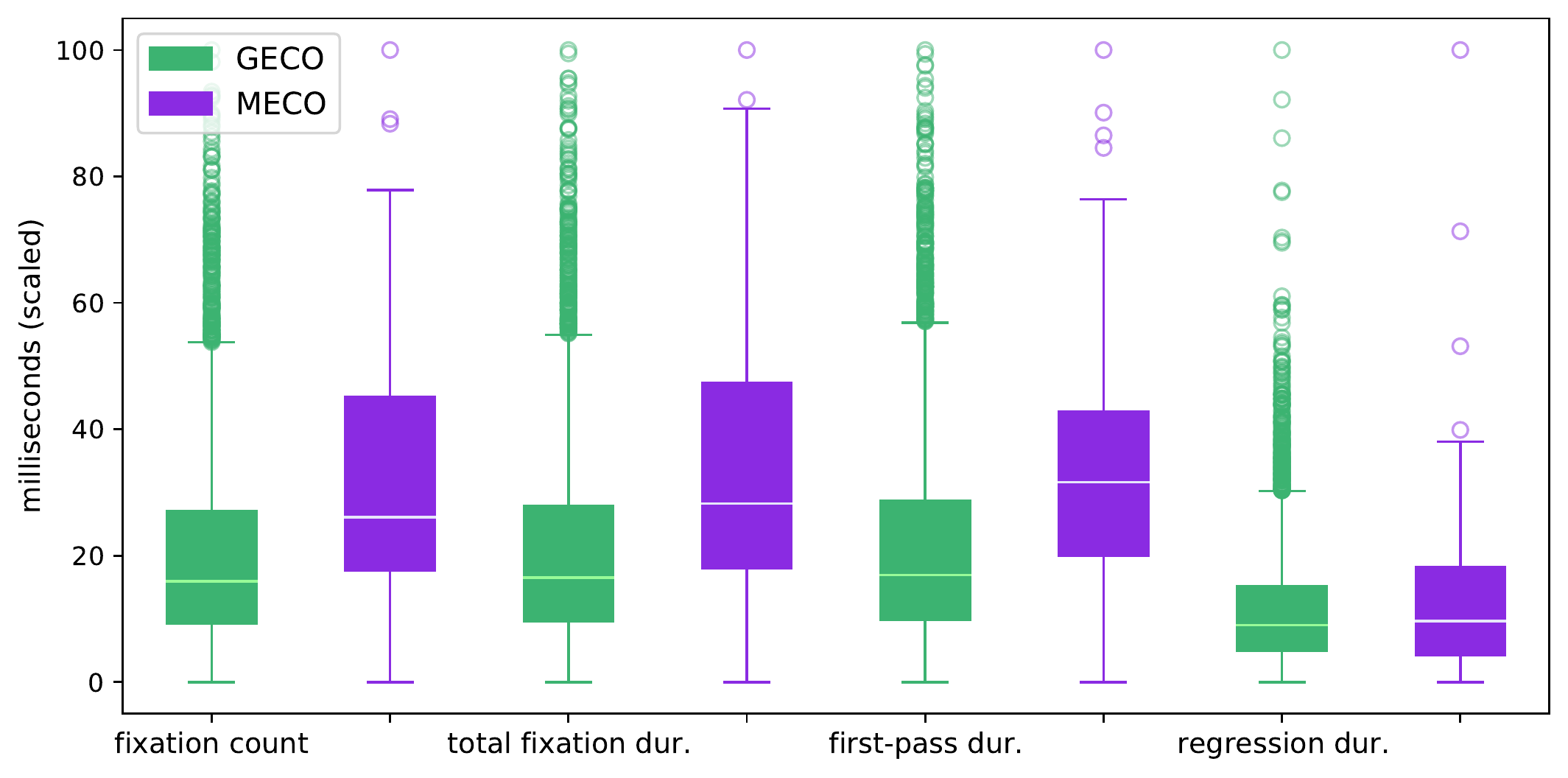}}
\caption{Distribution of four sentence-level eye-tracking metrics in English parts of GECO and MECO. All metrics are scaled between 0-100.
}
\label{fig:distribution-geco-meco}
\end{figure}

\begin{figure}[h!]
    \centering
    \includegraphics[scale=0.45]{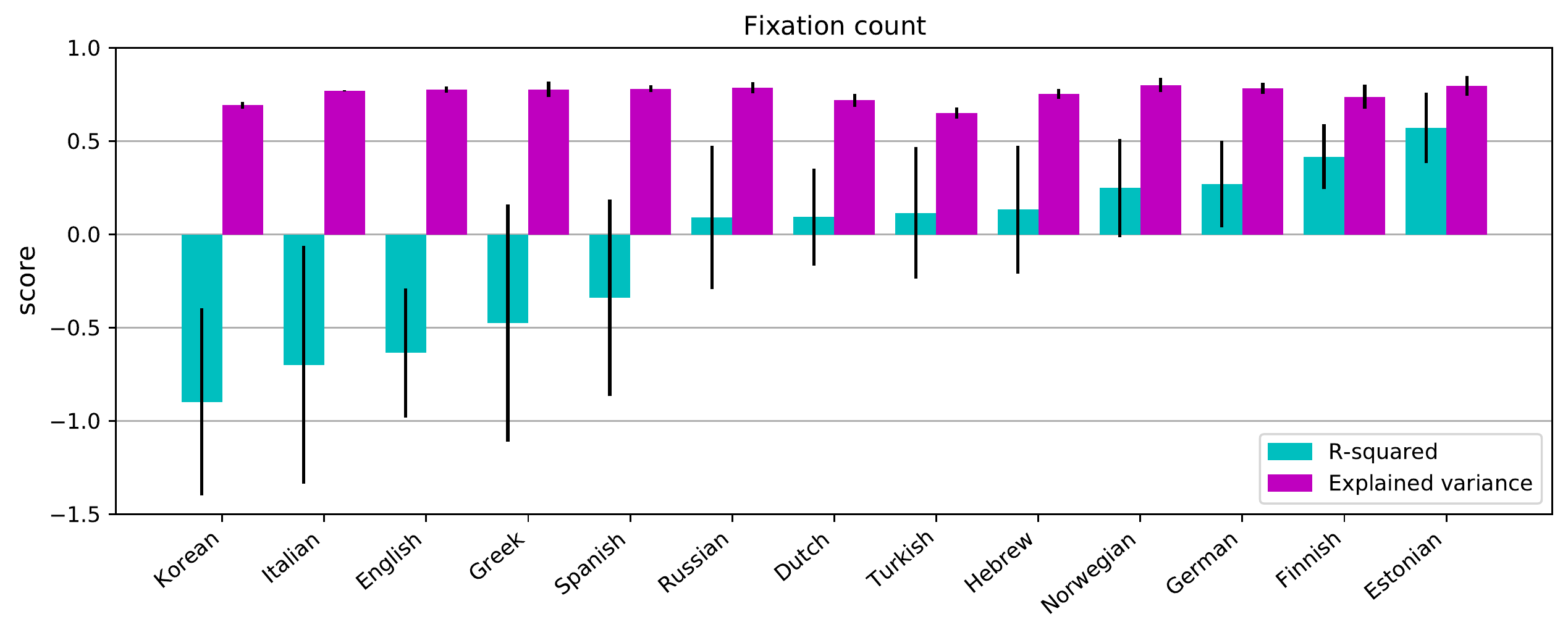}
    \includegraphics[scale=0.45]{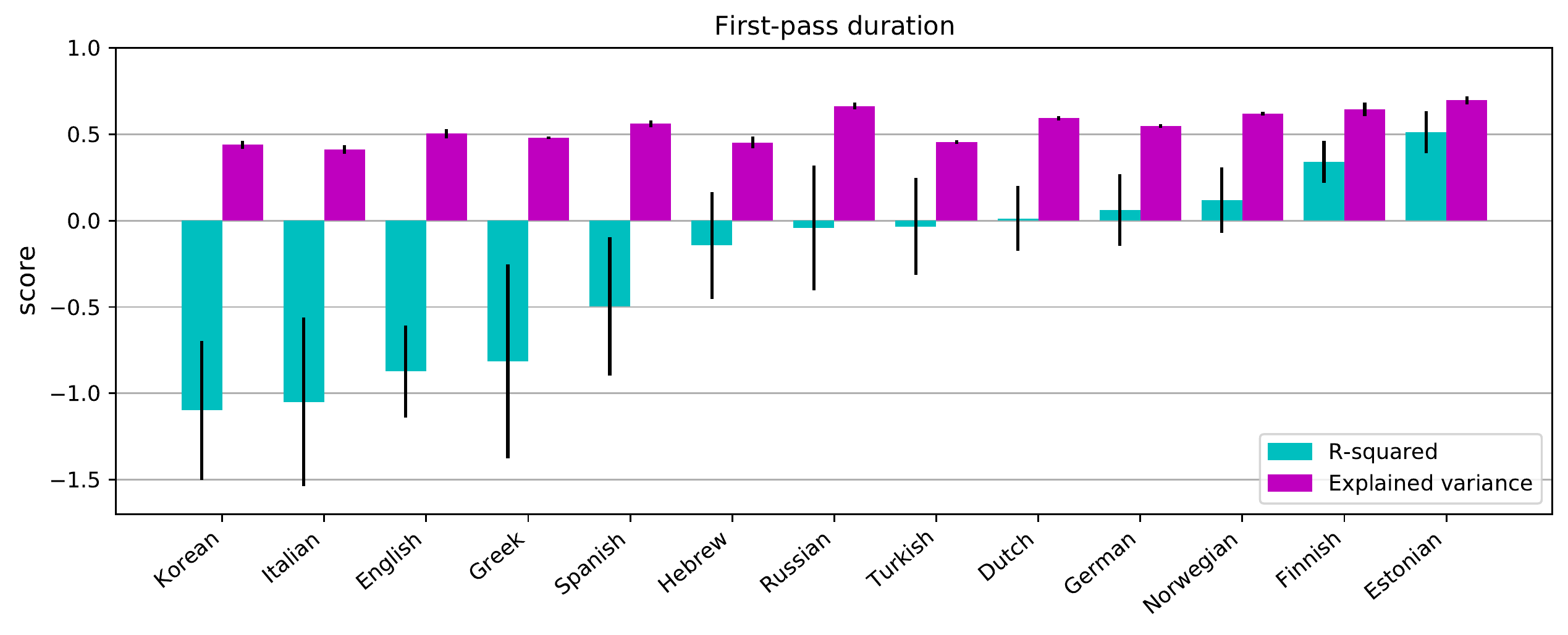}
    \includegraphics[scale=0.45]{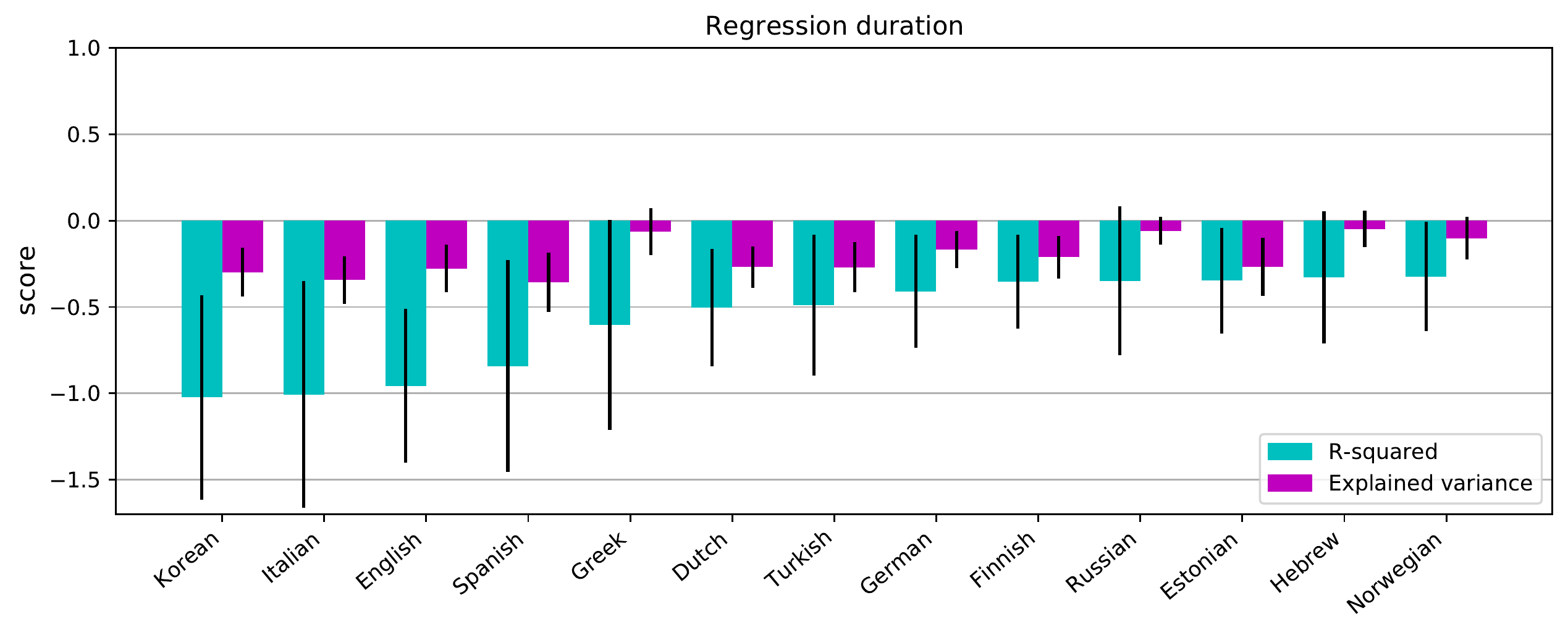}
    \caption{Cross-lingual transfer results for predicting cognitive processing complexity (i.e. fixation count, first-pass duration and regression duration). Prediction performance is evaluated with explained variance and $R^{2}$ for each language in MECO. The results are averaged over 5 folds; error bars denote the standard deviation over folds.}
    \label{crosslingual-accuracy-appendix}
\end{figure}

\begin{figure}[h!]
\centerline{\includegraphics[scale=0.5]{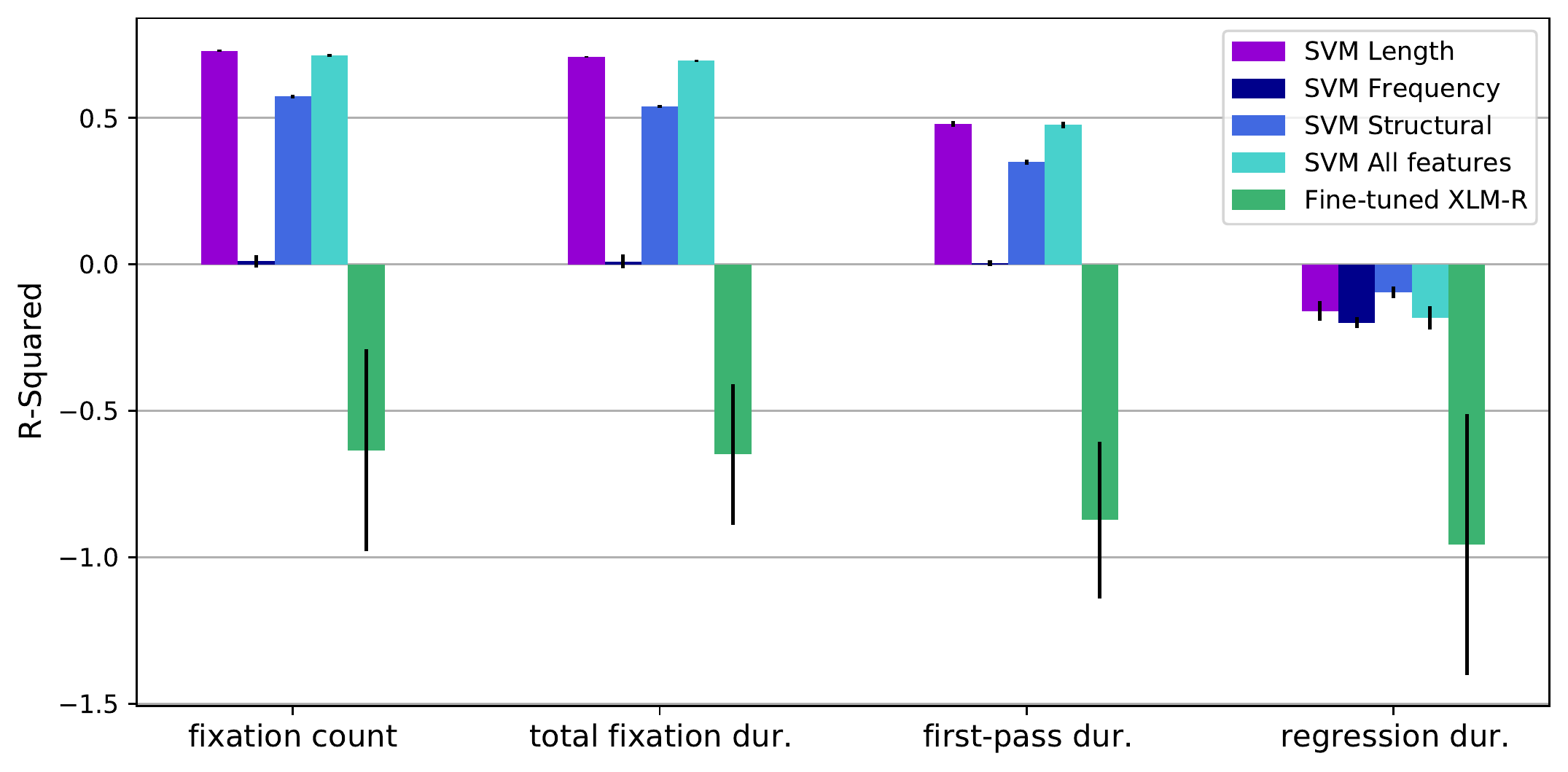}}
\caption{$R^{2}$ of the four feature-based SVM models and the fine-tuned XLM-R model. The models are trained on GECO using 5-fold cross-validation and evaluated on the English part of MECO; error bars denote the standard deviation over folds.
}
\label{fig:feature-based-prediction-r2}
\end{figure}

\begin{figure}[h!]
    \centering
    \includegraphics[scale=0.45]{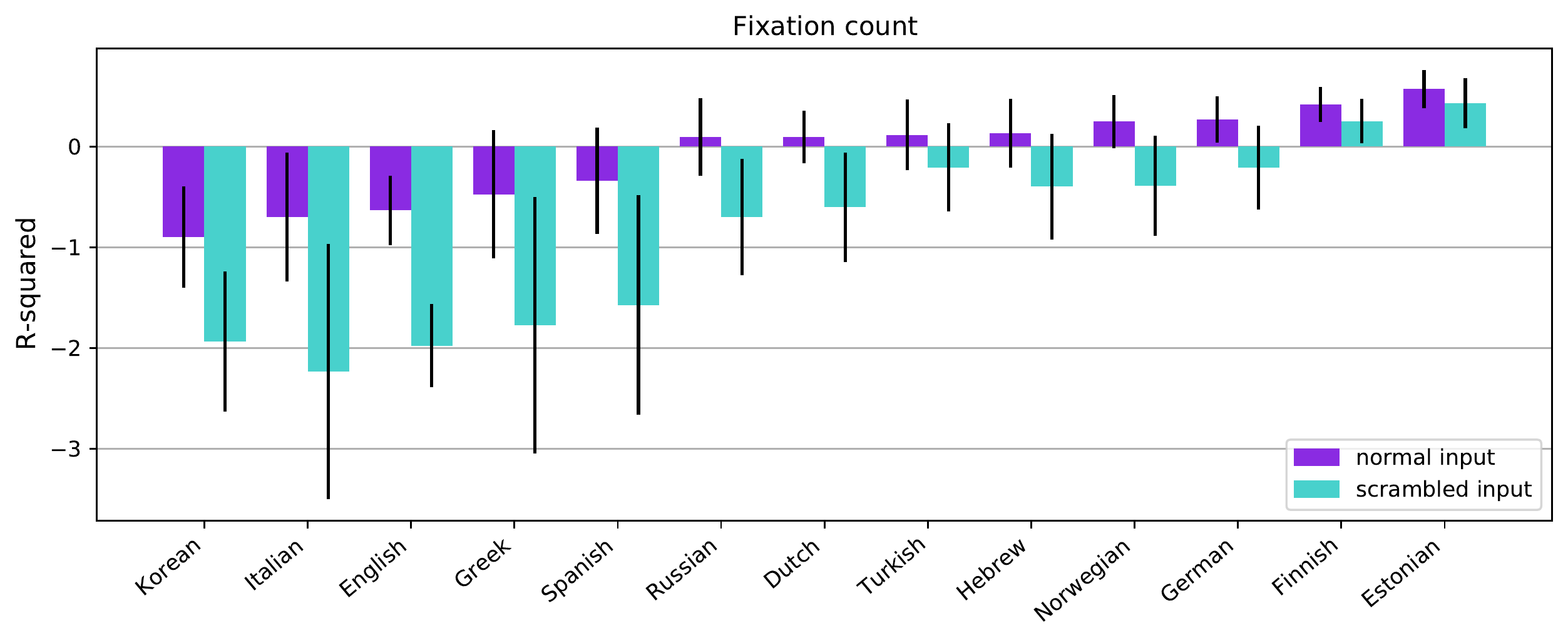}
    \includegraphics[scale=0.45]{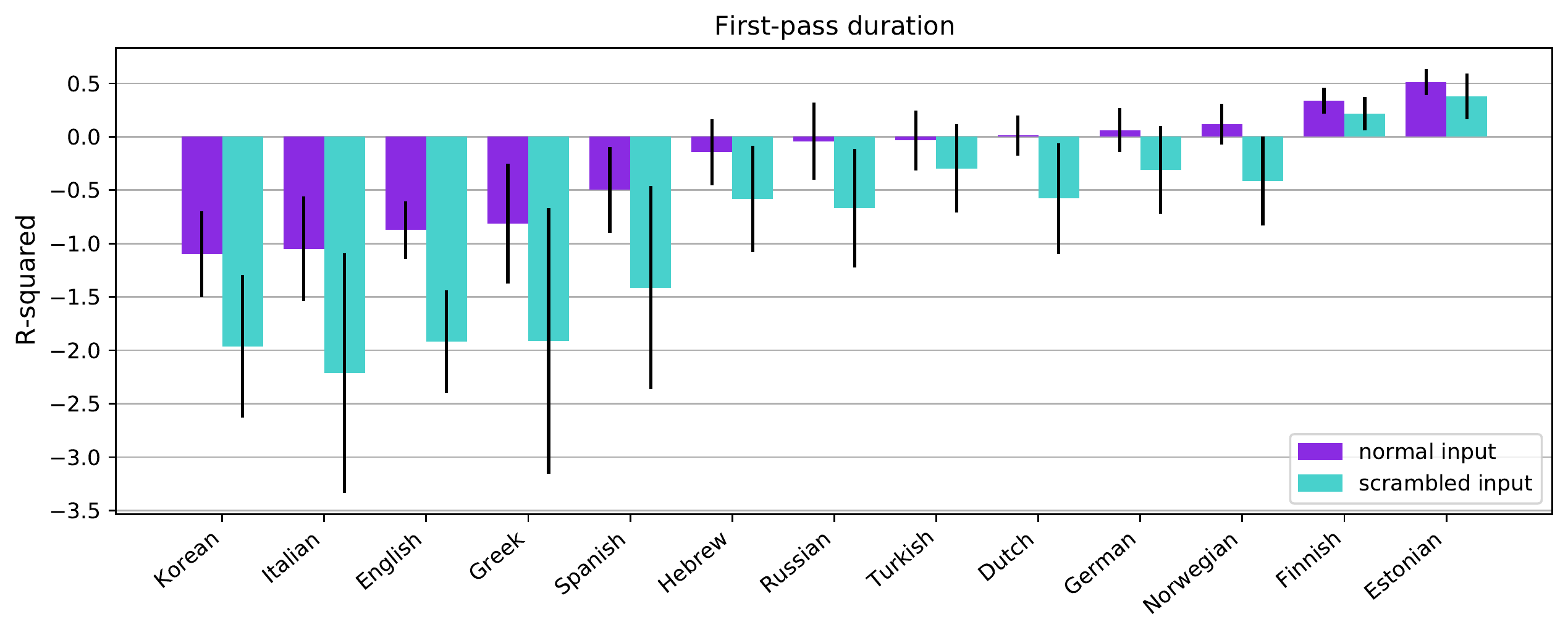}
    \includegraphics[scale=0.45]{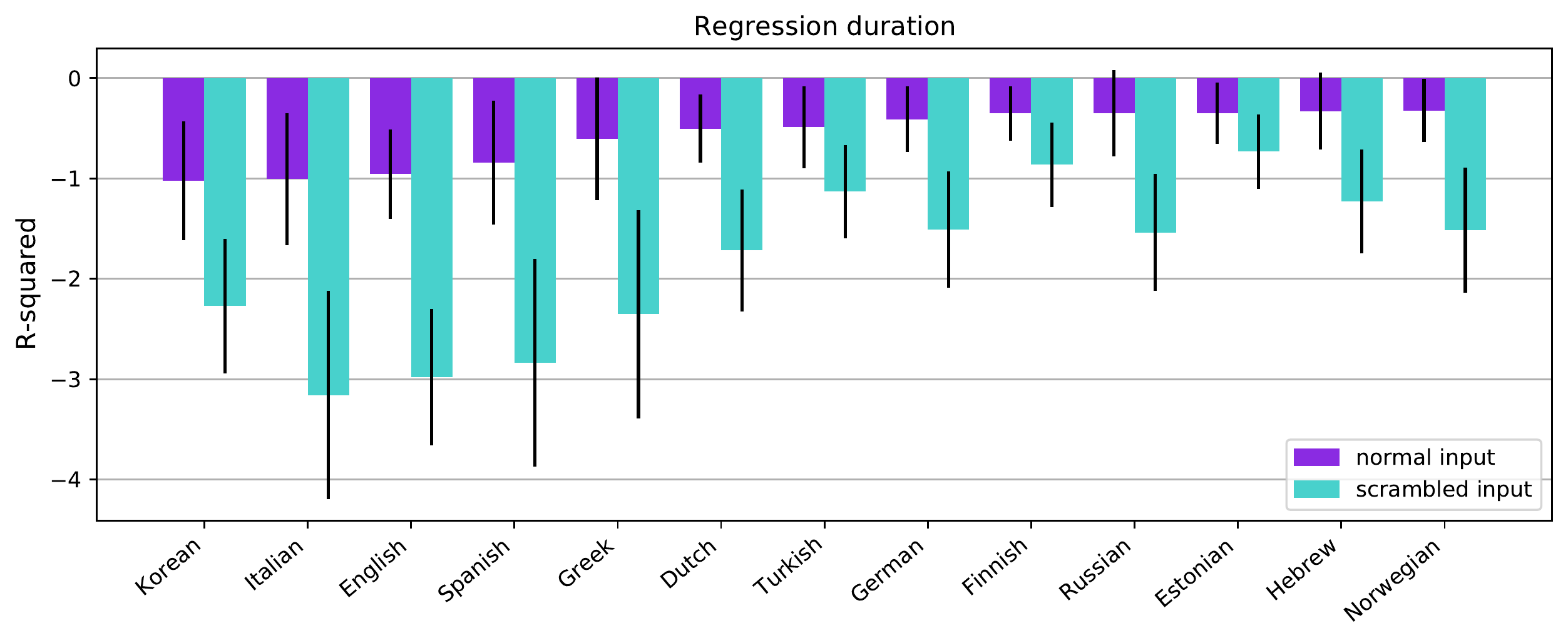}
    \caption{$R^{2}$ for fixation count, first-pass duration and regression duration for each language in MECO, both for sentences with normal and scrambled word order. The results are averaged over 5 folds; error bars denote the standard deviation.}
    \label{scrambled-results-appendix}
\end{figure}

\end{document}